\documentclass[sigconf]{acmart}
\usepackage{algorithm}
\usepackage{algorithmic}
\usepackage{balance}
\usepackage{amsfonts}
\usepackage{multirow}
\usepackage{subcaption}
\usepackage{stfloats}  

\AtBeginDocument{%
  }

\setcopyright{acmlicensed}

\copyrightyear{2026}
\acmYear{2026}
\setcopyright{cc}
\setcctype{by}
\acmConference[SAC '26]{The 41st ACM/SIGAPP Symposium on Applied Computing}{March 23--27, 2026}{Thessaloniki, Greece}
\acmBooktitle{The 41st ACM/SIGAPP Symposium on Applied Computing (SAC '26), March 23--27, 2026, Thessaloniki, Greece}
\acmPrice{}
\acmDOI{10.1145/3748522.3779746}
\acmISBN{979-8-4007-2294-3/2026/03}

\begin{document}

\title[\textsc{Pace}: Physics-Aware Attentive Temporal Convolutional Network]{\textsc{Pace}: Physics-Aware Attentive Temporal Convolutional Network \\for Battery Health Estimation}
\renewcommand{\shorttitle}{}

\makeatletter
\def\@affiliationfont{\fontsize{8.5pt}{11pt}\selectfont}  
\makeatother

\author{Sara Sameer}
\affiliation{%
  \institution{Singapore Institute of Technology}
  \country{Singapore}}
\email{sara.sameer@singaporetech.edu.sg}

\author{Wei Zhang}
\authornote{Corresponding author.}
\affiliation{%
  \institution{Singapore Institute of Technology}
  \country{Singapore}}
\email{wei.zhang@singaporetech.edu.sg}

\author{Dhivya Dharshini Kannan}
\affiliation{%
  \institution{Singapore Institute of Technology}
  \country{Singapore}}
\email{dhivyadharshini.kannan@singaporetech.edu.sg}

\author{Xin Lou}
\affiliation{%
  \institution{Singapore Institute of Technology}
  \country{Singapore}}
\email{lou.xin@singaporetech.edu.sg}

\author{Yulin Gao}
\affiliation{%
  \institution{ST Engineering}
  \country{Singapore}}
\email{gao.yulin@stengg.com}

\author{Terence	Goh}
\affiliation{%
  \institution{ST Engineering}
  \country{Singapore}}
\email{terence.goh@nus.edu.sg}

\author{Qingyu Yan}
\affiliation{%
  \institution{Nanyang Technological University}
  \country{Singapore}}
\email{alexyan@ntu.edu.sg}

\renewcommand{\shortauthors}{}

\begin{abstract}
Batteries are critical components in modern energy systems such as electric vehicles and power grid energy storage. Effective battery health management is essential for battery system safety, cost-efficiency, and sustainability. In this paper, we propose \textsc{Pace}\footnote{Code: https://github.com/SaraaSameer/PACE-for-Battery-Health-Estimation.}, a \underline{p}hysics-aware \underline{a}ttentive temporal \underline{c}onvolutional network for battery health \underline{e}stimation. \textsc{Pace} integrates raw sensor measurements with battery physics features derived from the equivalent circuit model. We develop three battery-specific modules, including dilated temporal blocks for efficient temporal encoding, chunked attention blocks for context modeling, and a dual-head output block for fusing short- and long-term battery degradation patterns. Together, the modules enable \textsc{Pace} to predict battery health accurately and efficiently in various battery usage conditions. In a large public dataset, \textsc{Pace} performs much better than existing models, achieving an average performance improvement of 6.5\% and 2.0x compared to two best-performing baseline models. We further demonstrate its practical viability with a real-time edge deployment on a Raspberry Pi. These results establish \textsc{Pace} as a practical and high-performance solution for battery health analytics.
\end{abstract}



\maketitle

\section{Introduction}
\label{sec:intro}
Batteries are the foundations for modern energy systems, such as electric vehicles (EVs) and energy storage of the power grid. As electrification accelerates worldwide, battery demand is rising rapidly. According to the IEA Global EV Outlook 2025 \cite{IEA25}, battery demand is largely driven by electric vehicles, e.g., lithium-ion batteries. The total demand for electric vehicles and energy storage reached 1 TWh in 2024, which is 25\% more than in 2023 and is projected to triple to over 3 TWh by 2030. This surge in demand highlights the importance of robust battery health management, not just to prevent unexpected failures but also to extend battery life and improve system dependability and cost efficiency. 

Battery degrades over time, and the process is inevitable and complex. The degradation is influenced by many aspects, such as electrochemical reactions, charging/discharging patterns, and driving behaviors \cite{9384220}. Among various indicators of battery health, the state of health (SoH) is one of the most commonly used metrics and reflects the battery’s current capacity relative to its new battery nominal capacity \cite{chen2023state}. Accurate SoH estimation is vital for battery safety, lifespan, and efficiency. Despite growing interest, developing SoH prediction models that are accurate and robust under diverse operating conditions remains a non-trivial challenge. This is especially true when models need to be lightweight for real-world deployment in battery systems with limited computing resources.

Traditional methods leverage domain knowledge to characterize battery behavior. For instance, the equivalent circuit model (ECM) represents a battery as a network of resistors and capacitors and is widely adopted for its simplicity and physical interpretability \cite{chen2023state}. Kalman filters are frequently applied to sensor measurements to estimate model parameters in real time \cite{11065210}. More detailed physics-based approaches, such as the pseudo–two-dimensional model \cite{9384220}, replicate lithium-ion transport and provide insights into mechanisms like solid-electrolyte interphase (SEI) growth and lithium plating. While these methods offer valuable physical intuition, they often rely on expert tuning, handcrafted features, and simplified electrochemical approximations, which can limit scalability and adaptability under diverse operating conditions.

Recently, data-driven and machine learning (ML) models have been proposed to learn complex degradation patterns directly from data using various algorithms. To model long sequences of battery data, methods such as long short-term memory networks (LSTMs) \cite{ouyang2024combined,eleftheriadis2024joint}, Transformers, temporal convolutional networks (TCNs) \cite{liu2024state, li2023makes}, and graph neural networks \cite{yao2023novel} have achieved competitive performance in battery analytics with their ability to capture long, nonlinear, and inconsistent behaviors of battery degradation. Hybrid models have also emerged to combine complementary strengths \cite{sameer2025ginet}. For example, incremental capacity analysis \cite{xu2024state} derives differential capacity curves from voltage–capacity data and adds them to ML feature spaces to improve accuracy. Physics-informed neural networks \cite{wang2024physics} have also been used to include physical constraints or governing equations into ML architectures or loss functions to improve prediction robustness, reduce data dependence, and enforce physically consistent behavior. However, because battery aging involves multiple coupled and nonlinear processes, formulating full partial differential equation models with tractable boundary conditions remains challenging in practice. In contrast, simpler empirical models like the ECM offer interpretable parameters without requiring full electrochemical modeling.

While existing ML models have achieved competitive performance on benchmark datasets, they often prioritize prediction accuracy over practical deployment. Many require substantial computing resources and struggle to generalize under varying deployment settings and operational conditions. For instance, Transformer-based SoH prediction models can exceed two million parameters, making them difficult to deploy in edge-based systems. In reality, battery management systems (BMS) operate under strict resource constraints, where SoH models must share limited computing power with other BMS functionalities. This combination of lightweight design, robustness, and high accuracy remains largely underexplored in the current literature.

In this paper, In this paper, we advance battery SoH prediction by combining domain knowledge with a tailored ML architecture. Our key idea is to jointly process physics-derived battery features and raw sensor data to enhance the modelling and learning of battery dynamics. Rather than simply concatenating features into generic ML models, we aim to develop specialized ML components to capture complex battery degradation patterns. Specifically, we propose \textsc{Pace}, a \underline{p}hysics-aware \underline{a}ttentive temporal \underline{c}onvolutional network for accurate and efficient battery SoH \underline{e}stimation. \textsc{Pace} integrates two complementary input modalities, including raw sensor readings such as voltage, current, and temperature, and physics-derived features such as internal resistance and open-circuit voltage (OCV) based on ECM. The model is composed of three purpose-built blocks. It comprises a temporal block for efficient encoding of long-range degradation trends, a streamlined attention block to capture fine-grained context, and a dual-head output block to extract both short- and long-term battery dynamics for accurate multi-scale prediction. Together, these blocks enable \textsc{Pace} to deliver high SoH prediction accuracy across diverse charging and discharging profiles while remaining compact and efficient for deployment in embedded BMS. We summarize our contributions as follows.
\begin{itemize}
    \item We propose \textsc{Pace}, a physics-aware deep learning model with battery-specialized modules that integrate sensor data and physics features for accurate and efficient SoH prediction.
    \item We conduct comprehensive experiments on a battery dataset under diverse operating conditions, and show that \textsc{Pace} consistently outperforms recent state-of-the-art models.
    \item We demonstrate the feasibility of deploying \textsc{Pace} at the edge by running it on a Raspberry Pi with real-time battery sensor readings. Our deployment showcases the practical viability of \textsc{Pace} for embedded BMS.
\end{itemize}

The rest of this paper is organized as follows. The next section presents battery physics, followed by the system architecture of \textsc{Pace} and technical details of the main components. Then we present experimental results, discussion, and a deployment demo. Finally, we conclude this paper and suggest future works.

\section{Battery Physics Modelling with ECM}
\label{sec:features}
A key strength of \textsc{Pace} is its usage of battery physics to complement raw sensor data. While common battery measurements such as voltage and current offer snapshots of battery dynamics, they lack the ability to capture internal battery electrochemical dynamics, which evolve with battery degradation. To better capture the battery's transient behavior, \textsc{Pace} extracts physics features from ECM and utilizes the features, compact yet expressive, for SoH prediction. Among various ECMs, we adopt the widely used first-order Thevenin model \cite{tran2021comparative} for its well-balanced trade-off between accuracy and efficiency. The model represents the battery as a combination of an ideal voltage source, internal resistance, and capacitance to reveal transient behavior with high modelling accuracy. The efficiency of the model allows us to inject physically meaningful features into \textsc{Pace} to achieve improved robustness and generalization without incurring significant computational overhead. Specifically, the first-order model consists of a few key components that collectively represent the battery's electrical behavior under various operating conditions \cite{tao2024state}. The first is the OCV, which is the battery's ideal voltage when no current is drawn. Next is the Ohmic resistance $R_0$, which models voltage's immediate drop when current is drawn, i.e., the battery's instantaneous response. The slow and transient changes in voltage are captured by a resistor-capacitor (RC) pair. It simulates the internal resistive and capacitive delays that prevent the battery from reaching expected voltage during charging or discharging. The model dynamics are described by the below equations.
\begin{equation}
V_{\text{out}} = V_0 - I R_0 - V_1,\quad
\overline{V_1} = -\frac{V_1}{R_1 C_1} + \frac{I}{C_1},
\end{equation}
where the battery's output voltage is $V_{\text{out}}$, which starts from OCV $V_0$, subtracts the immediate drop $I R_0$ with current $I$, and also subtracts transient voltage drop $V_1$ across the RC pair. The second equation shows the voltage changes across the RC pair, where $R_1$ and $C_1$ are the first order of polarization resistance and capacitance, respectively. The negative and positive terms represent gradual voltage decay and RC pair recharge by current input, respectively. Based on the equations, the foundational task of the model becomes estimating the parameters, including $V_0$, $R_0$, $R_1$, and $C_1$, to accurately model the dynamics of a specific battery. Parameters were identified by fitting the model to measured voltage–time data under applied current profiles using nonlinear least squares optimization. This approach adopts an offline baseline method to consistently extract physically meaningful parameters that capture battery resistance growth and dynamic polarization effects over aging. Future work can extend this approach to online and causal estimation methods such as Kalman filtering and recursive least squares.

\begin{figure}[t]
    \centering
    \hspace*{-0.025\linewidth}
    \includegraphics[width=1\linewidth]{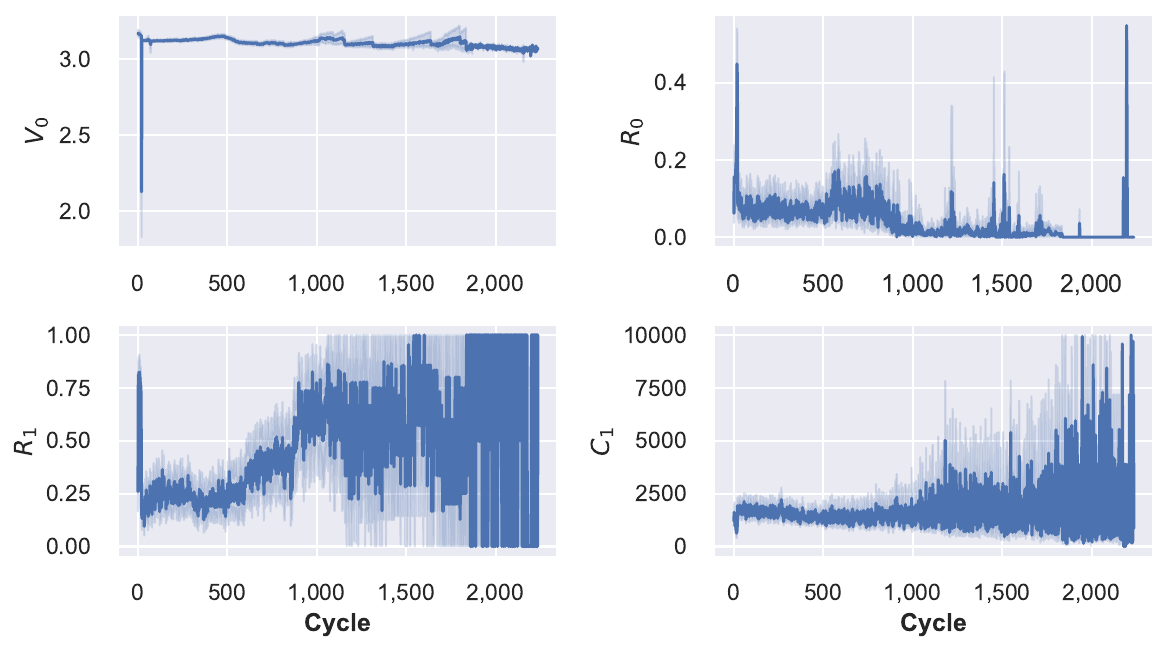}
    \caption{Evolution of ECM parameters throughout cycles.}
    \label{fig:ecm_parameter}
\end{figure}

Figure \ref{fig:ecm_parameter} presents the evolution of the estimated model parameters for a specific battery over its lifetime. We can see that OCV remains around 3.1 V with slow decay. This indicates that the battery maintains electrochemical (thermodynamic) equilibrium, and the charge-storage reactions remain stable and reversible before the battery degrades to its end-of-life (EoL). $R_0$ stabilizes rapidly after the initial spike, and the impact of Ohmic resistance on the battery's long-term degradation is relatively minimal. Differently, RC pair parameters change more and more significantly over the battery's lifetime and may indicate battery degradation due to reasons such as SEI growth and slower ion diffusion. Altogether, we use eight features, including voltage, current, temperature, cycle identifier, and the four physics features. Incorporating these lightweight yet physically grounded ECM features helps \textsc{PACE} track degradation trends more reliably and make robust and accurate predictions.

\begin{figure*}[]
    \centering
    \hspace*{-0.025\linewidth}
    \includegraphics[width=1.05\linewidth]{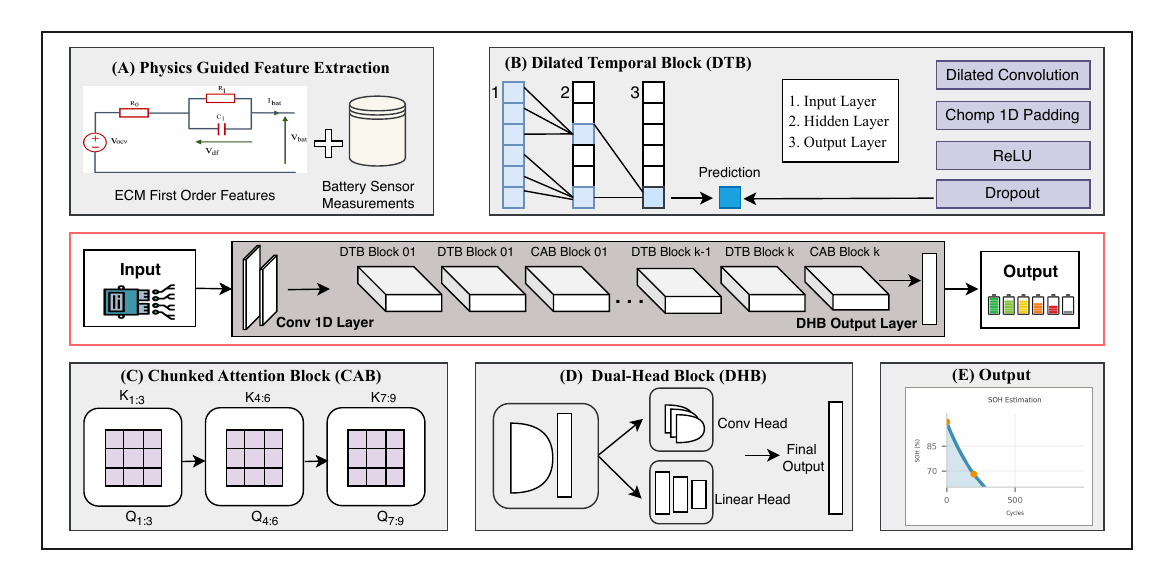}
    \caption{Our proposed design comprises two interconnected blocks, the dilated temporal block and the chunked attention block, followed by a dual-head output block. The input space contains the raw battery dataset's characteristics as well as the parameters retrieved by the ECM. The output layer has two heads and is responsible for multi-step predictions.}
    \label{fig:architecture}
\end{figure*} 

\section{\textsc{Pace}:  Methodology}
\label{sec:method}
To model battery SoH effectively, \textsc{Pace} combines physical insights with efficient temporal learning. As shown in Figure \ref{fig:architecture}, \textsc{Pace} is designed based on four tightly connected modules. It starts by obtaining battery sensor readings and extracting physics features based on ECM to reflect the battery's internal dynamics. Then, \textsc{Pace} captures battery degradation patterns with temporal convolution in dilated temporal block (DTB) and lightweight attention in chunked attention block (CAB). Several DTBs and CABs are interleaved as a sequence to enhance the modelling capability. Finally, \textsc{Pace} outputs the predicted SoH through a dual-head block (DHB). \textsc{Pace} is an integrated solution that is suited for real-world battery health modelling where accuracy, robustness, and deployability are essential. The details of the different modules of \textsc{Pace} are as below.

\subsection{Dilated Temporal Block (DTB)}
\label{sec:dtb}
Battery degradation is progressive and non-linear over time. To effectively learn the temporal dependencies embedded in both the ECM-derived features and raw data, we place the DTB immediately after the input to capture long-range patterns early in the pipeline of \textsc{Pace}, as shown in Figure \ref{fig:architecture}(B). DTB is used for sequential modelling based on dilated causal convolutions \cite{terzic2023tcnca, lai2023lightcts}. We follow the general structure of DTB and introduce battery-specific customizations detailed as follows. 

Instead of using the default exponential dilation scheme, we configure each dilated layer as a modular unit, consisting of a dilated convolution, \texttt{Chomp1D}, \texttt{ReLU}, and dropout. Our modular formulation offers greater flexibility and simplifies debugging. Its enhanced transparency makes it important for downstream operations in the industry. It also allows us to fine-tune each unit to the degradation characteristics of the battery. In typical implementations, dilated convolutions stack layers to expand the receptive field, which defines how far back in time a model can gather battery information for SoH prediction. We improve the receptive field configuration by aligning dilation rates and depth to the typical timescale of battery degradation, i.e., to ensure long-range modelling without excessive layer stacking. We further refine the residual formulation to enable deeper layers to adapt without compromising earlier battery trends. Unlike memory-intensive models, this design enables highly parallelized training and inference, making it ideal for embedded systems like BMS. In \texttt{Chomp1D}, we remove the excess padding introduced during convolution to maintain causal alignment and ensure that the SoH prediction at time $t$ does not include future padded inputs after time $t$. Finally, we apply normalization together with dropout to improve robustness to local fluctuations such as charging/discharging noise.

\begin{figure}[]
    \centering
    \hspace*{-0.075\linewidth}
    \includegraphics[width=1.15\linewidth]{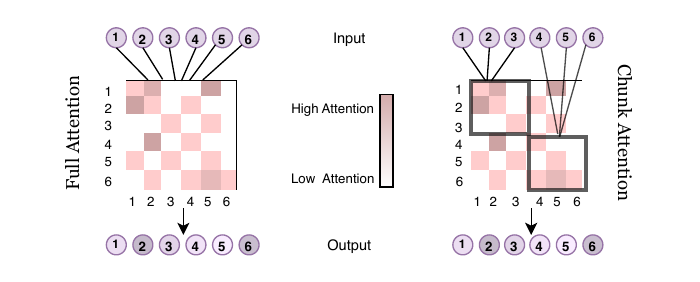}
    \caption{Comparison of attention mechanisms. Full attention computes dense token-to-token interactions in $ \mathcal{O}(n^2)$ time, and chunked attention limits computation to local sequence segments with $ \mathcal{O}(n)$ time complexity.}
    \label{fig:attention}
\end{figure} 

\subsection{Chunked Attention Block (CAB)}
DTB alone, based on local convolution, cannot well capture gradual and non-local patterns over the battery degradation horizons. While it captures structured dependencies through expanding receptive fields, it lacks the flexibility to adjust the weights or focus on different parts of the history data as degradation evolves. This motivates the use of a scalable attention mechanism to incorporate additional context. Full attention is often the default choice; however, it can be computationally expensive for long time-series sequences with its quadratic complexity $\mathcal{O}(n^2)$ with sequence length $n$. To incorporate scalable attention within temporal modeling, we introduced CAB \cite{zeineldeen2024chunked} into the pipeline, as presented in Figure \ref{fig:architecture}(C). 

CAB divides the input sequence into fixed-length chunks and applies attention independently within each non-overlapping temporal chunk. Figure \ref{fig:attention} uses a 6$\times$6 input sequence to demonstrate how attention weights are computed. In the case of full-attention, each output token focuses on all six input tokens, whereas CAB divides the input into two non-overlapping chunks of size 3 to restrict each output to attend only within the local chunk. As such, its complexity can be reduced to $\mathcal{O}(n)$. Furthermore, we optimize the placement of CAB in the pipeline of \textsc{Pace}. We interleave one CAB after every two DTBs to alternate between structured and convolutional memory, and attention-based recalibration. We repeat this DTB-DTB-CAB pattern several times. This strategy allows \textsc{Pace} to progressively build temporal features with increasing receptive fields and periodically calibrate its focus across sequence segments. Overall, we expect this design to enhance \textsc{Pace} to capture both fine-grained local dependencies and long-term degradation trends effectively and efficiently.

\subsection{Multi-Scale Dual-Head Output Block (DHB)}
Degradation evolves nonlinearly across a battery's lifecycle. It often starts slowly for new batteries and accelerates towards the batteries' EoL. Among different batteries, the capacity decay may show sharp drops, plateaus, or irregularities due to thermal, load, or manufacturing variations. Accurately predicting SoH thus requires the model to comprehensively consider degradations in different stages and battery settings. This should be addressed not only in hidden layers, e.g., in DTB and CAB, but also in output layers. In many predictive models, only a single linear decoder is configured in the output layer to map the extracted features from the hidden layers to the output. The simple design often struggles to generalize across different degradation patterns, e.g., over-smooth rapid decay and overfit to local noises. To address the challenges, we adopt a multi-head concept \cite{zhou2021informer} and propose DHB with two complementary heads, as depicted in Figure \ref{fig:architecture}(D). The first head is convolutional-based. It applies a 1D convolution over a fixed-size window to capture local and fine-grained temporal SoH dynamics. It also has a linear head, which configures a simple feedforward layer to capture global or smoothed trends and map the latent state to SoH prediction. The outputs of the two heads are fused into one, and the weight of each output is controlled by a learnable scalar gate $\alpha \in [0, 1]$. Let $\hat{y}_{\text{conv}}$ and $\hat{y}_{\text{linear}}$ be the outputs of the convolutional and linear heads, respectively. The predicted SoH is computed as,
\begin{equation}
\label{eq:output}
\hat{y}_{\text{SoH}} = \alpha \cdot \hat{y}_{\text{conv}} + (1 - \alpha) \cdot \hat{y}_{\text{linear}}.
\end{equation}

In summary, \textsc{Pace} is a physics-aware model for SoH prediction by integrating raw sensor measurements with physics-derived features from ECM to enrich the input representation. It combines DTB for temporal modelling, CAB for scalable context refinement, and DHB to fuse local and global degradation patterns to achieve accurate and generalizable battery health predictions.

\section{Experimental Study}
\label{sec:exp}
We conduct an experimental study in this section to evaluate the prediction performance of \textsc{Pace} comprehensively. The subsections provide details on the model configuration, hardware configurations, hyperparameter settings, and evaluation metrics. Following this, we also present a series of sensitivity analysis to investigate and quantify the individual contribution of physics-guided features and the chunk attention module to highlight their impact on the performance of the model.

\subsection{Experimental Setup}
We describe here the experimental setup, including the datasets employed and the model configuration with its training settings.

\subsubsection{Dataset}
We choose one of the largest public battery datasets from \cite{severson2019data} for our study. The dataset consists of 124 commercial Li-ion battery cells cycled under various charging/discharging conditions with diverse degradation patterns. Each cell has a nominal capacity of 1.1 Ah and was tested in a temperature-controlled chamber (30°C) until it reached 80\% of its initial capacity, i.e., EoL. The lifetime of the batteries varies between 150 and 2,300 cycles. Each cycle includes time-series sensor measurements for voltage, current, and temperature. SoH at cycle $t$ is given by $y_\text{SoH}^t = (Q_t / Q_0) \times 100\%$, where $Q_t$ and $Q_0$ are the maximum available capacity at cycle $t$ and initial capacity before cycling. SoH is modeled as a percentage, which shall not exceed 100\%, e.g., optimal condition for new batteries. To ensure a realistic evaluation, we follow the latest settings of \cite{xie2025data}, where 42 cells across 18 distinct charging protocols are carefully selected and used for training, and 6 cells with different charging conditions are selected for testing. Such a controlled configuration of training and testing data enables consistent and fair performance comparison.

\subsubsection{Model Configuration and Training Setup}
\textsc{Pace} is designed for multivariate and multi-step time-series prediction of SoH. We structure each batch of input sample as a tensor $\mathbf{X} \in \mathbb{R}^{B \times W \times F}$, where $B$ is the batch size, $W$ is the input window length, and $F$ is the number of features combining raw sensor measurements and ECM-derived physics features. We set $W$ to 100, which empirically can provide sufficient degradation dynamics while maintaining computational efficiency. The input first passes through two DTBs, each using a kernel size of 3 and an exponentially growing dilation factor $d_i = 2^i$ for layer $i$ of the block to rapidly expand the receptive field. CAB divides the temporal data into non-overlapping chunks of size 16 and applies multi-head self-attention to the chunks. We configure 8 attention heads, each operating on a distinct subset of feature dimensions, and attention is computed within each chunk to capture relatively local degradation patterns, without the quadratic cost of full-sequence attention. Altogether, the \textsc{Pace} pipeline has 6 DTBs and 3 CABs, each interleaved after every 2 DTBs. The output from the last CAB is passed to the only DHB in \textsc{Pace}. To fuse the outputs $\hat{y}_{\text{conv}}$ and $\hat{y}_{\text{linear}}$, we initialize $\alpha$ as 0.5, and it is learnable during model training, where \texttt{mse} is adopted as the loss function. 

We ran \textsc{Pace} with various hyper-parameters to evaluate their effect on performance and efficiency. We varied the channel widths of the temporal convolutional blocks, e.g., 32, 64, and 128, to assess their impact on performance. Standard additive residual connections were used in each block to stabilize training and accommodate more complex structures. Normalization was implemented as \texttt{LayerNorm} after attention modules to improve gradient flow and generalization, while the TCN blocks used weight normalization for faster convergence. Dropout rates were set to 0.2 for convolutional layers and 0.1 for the attention matrix to reduce overfitting. We trained with the \texttt{Adam} optimizer, a batch size of 32, and up to 200 epochs with early stopping (20-epoch patience) to prevent overfitting. Each experiment was repeated three times with different fixed seeds to ensure deterministic execution and reproducibility of results.

\subsubsection{Metrics}
\textsc{Pace} is designed with real-world deployment in mind. Instead of focusing solely on accuracy, as many academic studies do, we follow applied settings and evaluate both prediction accuracy and model efficiency, which is a key requirement for edge devices in battery health monitoring. Specifically, we not only adopt commonly used accuracy metrics, including mean absolute error (MAE) and root mean squared error (RMSE), but also quantify efficiency through model size and FLOPs. In addition, we are inspired by a recent work \cite{zhao2023deep} and define a metric $\eta$ called \textit{efficiency} as,
\begin{equation}
    \label{eq:efficiency}
     \eta\big(f(\cdot)\big) = \frac{1,000}{\text{RMSE}_f \times n_f},
\end{equation}
where $f(\cdot)$ is a prediction model, $\text{RMSE}_f$ is the RMSE achieved by the model, and $n_f$ is the number of parameters of the model. The metric measures the model's predictive performance per thousand parameters, based on the 1,000 in the denominator. Efficiency is different from the commonly used FLOPs, which means the floating point operations per second for estimating computational cost. Because FLOPs alone do not account for predictive accuracy or model size, our efficiency metric is more suitable for assessing deployment readiness.

\begin{table*}[]
\centering
\caption{Performance comparison across models on battery SoH prediction. Metrics include the number of model parameters, FLOPs, RMSE, MAE, efficiency $\eta$ (higher $\eta$ indicates better accuracy per model size), the mean $\eta$ across prediction horizons, and the relative efficiency ratio (normalized to the best-performing model). $\uparrow$ ($\downarrow$) indicate that higher (lower) is better. Best results are highlighted in bold, and \textsc{Pace} achieves the highest efficiency.}
\def\arraystretch{1.2}
\small

\resizebox{\textwidth}{!}{ \begin{tabular}{c|rr|rrr|rrr|rrr|rr}
\hline\hline
\multirow{2}{*}{Model}
& \multicolumn{1}{c}{\#Params $\downarrow$} & \multicolumn{1}{c|}{FLOPs $\downarrow$} & \multicolumn{3}{c|}{1-Cycle} & \multicolumn{3}{c|}{30-Cycle} & \multicolumn{3}{c|}{50-Cycle} & \multicolumn{2}{c}{$\eta$} \\ \cline{4-14}
& \multicolumn{1}{c}{($\times 10^3$)} & \multicolumn{1}{c|}{($\times 10^6$)} & \multicolumn{1}{c}{RMSE $\downarrow$} & \multicolumn{1}{c}{MAE $\downarrow$} & \multicolumn{1}{c|}{$\eta$ $\uparrow$} & \multicolumn{1}{c}{RMSE $\downarrow$} & \multicolumn{1}{c}{MAE $\downarrow$} & \multicolumn{1}{c|}{$\eta$ $\uparrow$} & \multicolumn{1}{c}{RMSE $\downarrow$} & \multicolumn{1}{c}{MAE $\downarrow$} & \multicolumn{1}{c|}{$\eta$ $\uparrow$} & \multicolumn{1}{c}{Mean $\uparrow$} & \multicolumn{1}{c}{Ratio $\uparrow$} \\
\hline\hline
Transformer & 2,559.5 & 133.2 & 0.014 & 0.009  & 27.9 & 0.016  & 0.010 & 24.4 & 0.017  & 0.009  & 23.0 & 25.1 & 5.2\% \\ \hline
TCN {} & 47.0 & 1.7 & 0.036  & 0.022  & 580.2 & 0.053  & 0.039 & 401.8 & 0.057  & 0.042  & 373.6 & 451.9 & 94.0\% \\ \hline
BCA & 173.5 & 5.0 & 0.034 & 0.018 & 169.5 & 0.037 & 0.020  & 155.8 & 0.038  & 0.020 & 150.5 & 158.6 & 33.0\% \\ \hline
\textbf{\textsc{Pace} (ours)} & 70.9 & 5.1 & 0.023 & 0.010  & \textbf{{613.4}} & 0.033 & 0.014 & \textbf{{427.5}} & 0.035  & 0.015 & \textbf{{403.1}} & \textbf{481.3} & \multicolumn{1}{c}{$-$} \\
\hline\hline
\end{tabular}
}
\label{tab:comp}
\end{table*}

\subsection{Comparison Study}
We present a comparison study in this part.

\subsubsection{Comparison Algorithms}
We compare \textsc{Pace} with several representative baselines, each of which reflects a different modeling philosophy. First, we choose a model that is accuracy-driven with no restrictions on model size. We expect our proposed \textsc{Pace} to approximate the accuracy of this model as much as possible and, meanwhile, maintain its advantage in practical edge deployment. Specifically, we choose Transformer \cite{sun2025optimizing}, which is attention-based and offers high prediction accuracy for SoH. However, it has significant computational overhead and is impractical for embedded battery systems. Second, we prioritize model size and choose TCN. It is a lightweight model based on dilated causal convolutions and known for its efficiency. We expect \textsc{Pace} to be as compact as possible in comparison with TCN and able to perform well for both short-term and long-term SoH prediction. Finally, we choose one of the latest models, BiLSTM-CNN-Attention \cite{xie2025data}, referred to as BCA for simplicity. The model has a hybrid design and combines recurrent, convolutional, and attention mechanisms. It is one of the most competitive models for SoH prediction based on our survey.

\subsubsection{Results}
Table \ref{tab:comp} shows the comparison results achieved by \textsc{Pace} and three comparison algorithms across different prediction horizons. Let us first compare \textsc{Pace} with Transformer. \textsc{Pace} achieves the best trade-off between prediction accuracy and model efficiency across all horizons. For 1-cycle prediction, it achieves the highest efficiency of 613, while maintaining strong performance even at 50 cycles with efficiency 403. Compared to Transformer, which has an extremely large model size, with 2.6M model parameters and 133M FLOPs, \textsc{Pace} is a significantly more practical solution for battery systems. Transformer achieves only 4.5\%, 5.7\%, and 5.7\% of \textsc{Pace}’s efficiency for 1-cycle, 30-cycle, and 50-cycle prediction horizons, respectively. The efficiency advantage of \textsc{Pace} is driven by its lightweight yet expressive design, with DTB for temporal patterns, CAB for contextual focus, and DHB for adaptive fusion of different degradation trends. Together, \textsc{Pace} approximates Transformer’s prediction accuracy well with only 2.8\%  of the model parameters and 3.8\% of the FLOPs of Transformer. 

The second comparison algorithm, TCN, achieves significantly better efficiency compared to Transformer, largely due to its lightweight convolutional design. It comes remarkably close to \textsc{Pace} and approximates the efficiency by 94.6\%, 94.0\%, and 92.7\% for 1-cycle, 30-cycle, and 50-cycle prediction horizons, respectively. For example, TCN achieves an efficiency score of 580 versus 613 for \textsc{Pace} for 1-cycle prediction, using only 47k parameters and 1.7M FLOPs. We notice that TCN approximates \textsc{Pace} relatively better for short-term prediction, which supports the argument that dilated causal convolution is more capable for short-term compared to long-term predictions. Unlike TCN, \textsc{Pace} has a holistic design with different modules for various degradation patterns and can generalize for SoH prediction in different timescales. 

\textsc{Pace} also outperforms BCA, one of the newest and most advanced hybrid models, for all tested horizons in terms of different performance metrics, e.g., RMSE, MAE, and efficiency. While BCA integrates several components such as convolution and attention to well capture temporal and spatial dynamics, its complexity does not yield optimal performance. For 50-cycle prediction, \textsc{Pace} achieves an MAE of 0.015 and an efficiency score of 403, which are 33\% and 2.7x better than BCA, respectively. This highlights the advantages of \textsc{Pace} and shows that a carefully designed lightweight model can outperform computationally intensive and hybrid models in both prediction accuracy and deployment efficiency.

\subsection{Impact of Physics}
\begin{figure}[t]
    \centering
    \hspace*{-0.16in}
\includegraphics[width=1.1\linewidth]{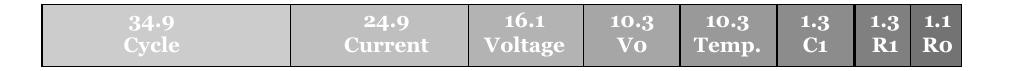}
    \caption{Feature importance (in percentage) distribution for the used features in \textsc{Pace}. The physics features complement the raw features for achieving optimal model performance.}
    \label{fig:importance}
\end{figure}

We now evaluate the contribution of physics-derived battery features to the overall effectiveness of \textsc{Pace}. Figure \ref{fig:importance} shows the feature importance distribution of the input features via permutation importance, each as a percentage, and the summation is 100\%. We can see that physics features do not dominate the prediction. However, they encode degradation trends that complement the noisy and high-frequency sensor readings and offer physically grounded signals that are less prone to overfitting.
To do this, we compare our full \textsc{Pace} model against its variant that excludes the four physics features and relies solely on raw sensor data. The core architecture, including DTB, CAB, and DHB, remains unchanged between the two models, except for minor customization to accommodate the reduced input dimensionality in the variant. 

\begin{table}[t]
\centering
\caption{Performance of \textsc{Pace} (the 1st row) and its variant without using physics (the 2nd row) across different prediction horizons. A small value indicates good performance. \textsc{Pace} with physics integrated has improved SoH prediction.}
\def\arraystretch{1.2}
\small
\setlength{\tabcolsep}{6pt}
\resizebox{\linewidth}{!}{ \begin{tabular}{c|cc|cc|cc}
\hline\hline
\multirow{2}{*}{Physics} & \multicolumn{2}{c|}{1-Cycle} & \multicolumn{2}{c|}{30-Cycle} & \multicolumn{2}{c}{{50-Cycle}} \\
\cline{2-7}
& {RMSE} & {MAE} & {RMSE} & {MAE} & {RMSE} & {MAE} \\
\hline\hline
$\checkmark$ & \textbf{0.023} & \textbf{0.010} & \textbf{0.033} & \textbf{0.014} & \textbf{0.035} & \textbf{0.015} \\\hline
$\times$           & 0.031 & 0.014 & 0.045 & 0.016 & 0.048 & 0.016 \\
\hline\hline
\end{tabular}
}
\label{tab:physics}
\end{table}

As shown in Table \ref{tab:physics}, the full \textsc{Pace} model outperforms its physics-agnostic variant across all tested prediction horizons in terms of both RMSE and MAE. Take the 50-cycle prediction, for example; the physics features help \textsc{Pace} reduce RMSE by 27\% from 0.048 to 0.035 compared to its variant. The performance improvement can be attributed to the value that the four ECM-derived physics features contribute. $V_0$ and $R_0$ reflect the equilibrium voltage and immediate resistive loss, respectively, and both tend to decline as the battery degrades. $R_1$ and $C_1$, representing the RC pair, capture slower transient electrochemical behaviour and are both sensitive to battery aging. These features offer a physically meaningful view of internal battery dynamics, and unfortunately, such physics information is difficult to infer directly from raw sensor measurements such as current and temperature using data-driven methods. Guided by such physically grounded features, \textsc{Pace} gains a comprehensive understanding of degradation and generalizes in various settings.

\begin{table*}[]
\centering
\caption{Comparison of attention mechanisms within \textsc{Pace} for battery SoH prediction. Metrics include the number of parameters, FLOPs, RMSE, MAE, efficiency $\eta$ (higher $\eta$ indicates better accuracy per model size), the mean $\eta$ across prediction horizons, and the relative efficiency ratio (normalized to the best-performing model). $\uparrow$ ($\downarrow$) indicate that higher (lower) is better. Best results are highlighted in bold, and \textsc{Pace} with CAB achieves the highest efficiency.}

\def\arraystretch{1.2}
\small 
\resizebox{\textwidth}{!}{ \begin{tabular}{c|cc|ccc|ccc|ccc|cc}
\hline\hline
\multirow{2}{*}{Attention}
& \multicolumn{1}{c}{\#Params $\downarrow$} & \multicolumn{1}{c|}{FLOPs $\downarrow$} & \multicolumn{3}{c|}{1-Cycle} & \multicolumn{3}{c|}{30-Cycle} & \multicolumn{3}{c|}{50-Cycle} & \multicolumn{2}{c}{$\eta$} \\ \cline{4-14}
& \multicolumn{1}{c}{($\times 10^3$)} & \multicolumn{1}{c|}{($\times 10^6$)} & \multicolumn{1}{c}{RMSE $\downarrow$} & \multicolumn{1}{c}{MAE $\downarrow$} & \multicolumn{1}{c|}{$\eta$ $\uparrow$} & \multicolumn{1}{c}{RMSE $\downarrow$} & \multicolumn{1}{c}{MAE $\downarrow$} & \multicolumn{1}{c|}{$\eta$ $\uparrow$} & \multicolumn{1}{c}{RMSE $\downarrow$} & \multicolumn{1}{c}{MAE $\downarrow$} & \multicolumn{1}{c|}{$\eta$ $\uparrow$} & \multicolumn{1}{c}{Mean $\uparrow$} & \multicolumn{1}{c}{Ratio $\uparrow$}\\
\hline\hline
CAB (ours) & 70.9 & 5.14  & 0.023 & 0.010 & \textbf{613.4} & 0.033 & 0.014 & \textbf{{427.5}} & 0.035 & 0.015 & \textbf{{403.1}} & 481.3 & \multicolumn{1}{c}{$-$} \\ \hline
Single-head & 66.7 & 5.14 & 0.027 & 0.013 & 555.1 & 0.036 & 0.016 & 416.3 & 0.038 & 0.016 & 394.4 & 455.3 & 94.6\% \\ \hline
Multi-head & 71.0 & 5.14 & 0.025 & 0.011 & 563.3 & 0.033 & 0.014 & 426.7 & 0.035 &  0.015 & 402.4 & 464.1 & 96.5\% \\
\hline\hline
\end{tabular}
}
\label{tab:attention}
\end{table*}

\subsection{Impact of Attention}
CAB is one of the building blocks of \textsc{Pace} for its unique attention mechanism. In this part, we aim to understand the role of attention configuration within our model and conduct a comparison of different attention designs used in \textsc{Pace}. For this, we compare the default \textsc{Pace} with CAB with two variants. One adopts the single-head attention that computes attention without parallel heads. The other is based on multi-head attention, i.e., using full attention without chunking as CAB does. The models are kept identical except for attention mechanisms, and the results are available in Table \ref{tab:attention}.

From the number of parameters and FLOPs, the differences across the three models appear marginal, e.g., a 6\% difference in model size between the models based on CAB and single-head attention, respectively. This is as expected, as attention is only one component within \textsc{Pace}, and reducing attention complexity affects only a subset of the model. However, these architectural choices still lead to meaningful differences in battery analytics accuracy and efficiency. 

Among the three variants, CAB outperforms single-head attention across all tested prediction horizons. For instance, CAB achieves an RMSE of 0.023 compared to 0.027 with single-head attention, about a 15\% reduction in error. Although single-head is relatively more lightweight, CAB still manages to perform well with both accuracy and efficiency being considered, and its efficiency score is 613, 10.5\% higher than 555 for the single-head attention. The performance advantage of CAB persists for long-term predictions as well. Relatively, the advantage of CAB is attributed not to model size but to its attention strategy. By dividing input sequences into chunks that are non-overlapping and computing attention locally, CAB captures degradation within one or a few cycles more effectively, while single-head attention struggles to capture important patterns comprehensively. 

CAB is smaller in size than multi-head attention, and it still achieves the same level of accuracy as the latter for the tested horizons, while using fewer parameters and less memory footprint on embedded hardware because its chunk-based computation avoids computing full attention across all tokens. For example, both achieve the same RMSE of 0.033 for 30-cycle prediction. CAB is even more accurate with 8\% RMSE improvement compared to multi-head attention for 1-cycle prediction. As such, CAB leads to higher overall efficiency, e.g., 9\% better than multi-head attention. These results show that CAB, as a lightweight chunk-based attention, can replace heavier attention to predict accurately while being resource-efficient and faster for battery-side deployment.

\subsection{Edge Deployment Demo}
To well demonstrate the practical utility of the proposed \textsc{Pace}, we deploy the model on an edge device for real-time battery health monitoring. Edge deployment is critical for industry scenarios such as EVs and distributed energy storage systems, where computing resources are limited and the requirements of real-time decision-making are high. Running \textsc{Pace} on a compact and energy-efficient platform proves its readiness for real-world applications. 

\begin{figure}[t]
    \centering	
    \hspace{-0.75em}
    \begin{subfigure}[c]{1.8in}
		\centering
		\includegraphics[width=\linewidth]{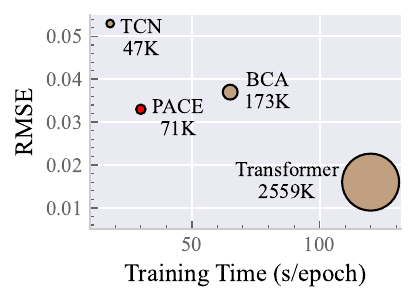}

    \end{subfigure}
    \hspace{-1em}
    \begin{subfigure}[c]{1.65in}
		\centering
		\includegraphics[width=\linewidth]{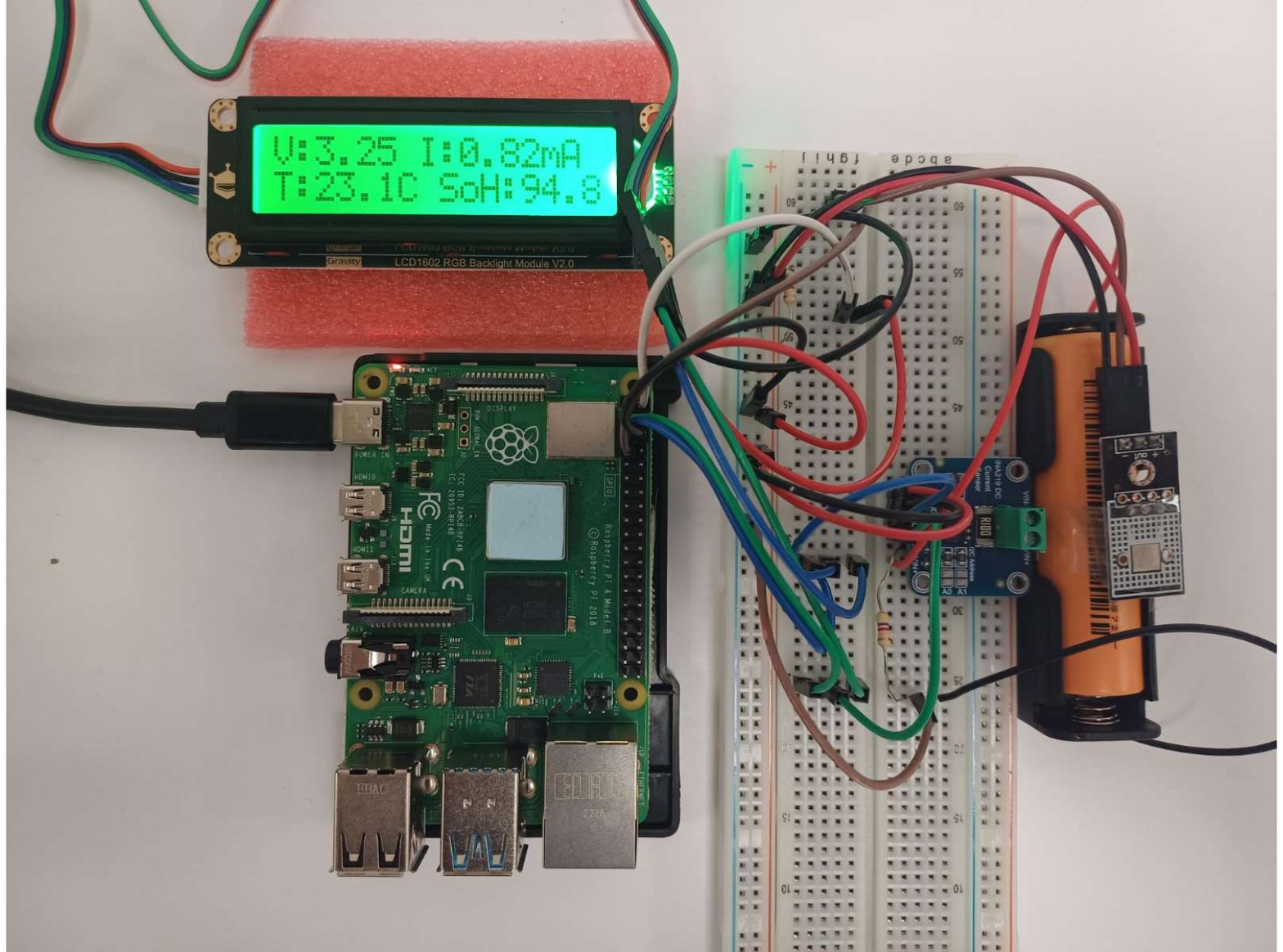}
                
    \end{subfigure}
    \caption{Left: Comparison of training time vs. RMSE across different models. Each dot represents a model, with dot size proportional to the number of parameters. \textsc{Pace} achieves a balance between deployability and accuracy. Right: Real-world deployment of \textsc{Pace} on a Raspberry Pi for real-time battery SoH monitoring. The demo showcases its lightweight and practical edge implementation.}
    \label{fig:rpi}
\end{figure}

Our demo of edge deployment is built around a Raspberry Pi 4, as shown in Figure \ref{fig:rpi}. The figure also illustrates the trade-off between training time and model accuracy. \textsc{Pace} achieves a favorable balance with $71$K parameters and a training time of approximately $40$ seconds per epoch, while performing comparably. For deployment, we integrate multiple sensors, including an INA219 module for real-time voltage and current measurements and a DS18B20 digital sensor for temperature. The former communicates via the Pi’s SDA pin, and the latter uses the 1-Wire protocol. Data from these sensors is first processed locally using Python libraries such as \texttt{w1thermsensor}, \texttt{smbus2}, and \texttt{ina219}. A Python script continuously computes the first-order ECM features from raw sensor data, and the features are passed into \textsc{Pace} that runs inference directly to estimate battery SoH. The key information is visualized in real-time on an I2C-enabled LCD1602 RGB Backlight Module V2.0, which communicates via the Pi’s SCL pin and has two I2C addresses for lighting control and display content change. In summary, our embedded solution runs all calculations locally without accessing remote servers, and the SoH can be estimated in real-time. We can see that the usage of physics improves the model’s accuracy with minimized computational overhead. The demo shows the potential of \textsc{Pace} for estimating SoH well in low-resource and real-time environments, and demonstrates the viability for real-world applications such as EV and power grid energy systems.

\balance
\section{Conclusion} 
\label{sec:conclusion}
In this paper, we propose \textsc{Pace}, a physics-aware deep learning framework for short-term and long-term battery health prediction. \textsc{Pace} bridges the gap between domain knowledge, which is highly important for industry applications, and data-driven learning. It integrates raw data from battery sensor measurements and physics features derived from an ECM. It also combines three battery-specific components, including DTB for efficient temporal representation, CAB for attention-based context modelling, and DHB for fusing diverse degradation dynamics. These components together enable \textsc{Pace} to capture both short-term and long-term battery aging patterns, without requiring extensive computing resources. We evaluate the performance of \textsc{Pace} on a large public dataset under different charging/discharging conditions. \textsc{Pace} shows consistent improvements over state-of-the-art comparison algorithms in terms of SoH prediction accuracy and model efficiency, achieving average efficiency improvements of 6.5\% and 2.0x compared to the two best-performing comparison algorithms, respectively, across varying prediction horizons. We demonstrate the critical roles of physics features and different \textsc{Pace} components in enhancing model performance. Overall, \textsc{Pace} is a practical model that is efficient and accurate and serves as a step toward robust and scalable battery analytics for diverse deployment scenarios, including real-time inference on edge devices, as we have demonstrated on a Raspberry Pi.

\begin{acks}
This research is supported by A*STAR under its MTC Programmatic (Award M23L9b0052), MTC Individual Research Grants (IRG) (Award M23M6c0113), SIT’s Ignition Grant (STEM) (Grant ID: IG (S) 2/2023 – 792), and the National Research Foundation Singapore and DSO National Laboratories under the AI Singapore Programme (AISG Award No: AISG2-GC-2023-006).
\end{acks}

\bibliographystyle{ACM-Reference-Format}
\bibliography{sample-base}
\end{document}